\documentclass[conference]{IEEEtran}
\usepackage{cite}
\usepackage{amsmath,amssymb,amsfonts}
\usepackage{algorithmic}
\usepackage{graphicx}
\usepackage{textcomp}

\usepackage[english]{babel}
\usepackage[T1]{fontenc}
\usepackage{epsfig}
\IEEEoverridecommandlockouts
\usepackage{tabu}
\usepackage{float}
\usepackage{amsmath}
\usepackage{amssymb}
\usepackage{amsthm}
\usepackage[keeplastbox]{flushend}
\usepackage{cite}
\usepackage{subfigure}
\usepackage{epstopdf}
\usepackage{multirow}
\usepackage{caption}
\usepackage[table]{xcolor}
\usepackage{mathtools}
\usepackage{xfrac}
\usepackage{units}
\usepackage[hyphens]{url}
\usepackage{array}
\newcommand{\beql}[1]{\begin{equation}\label{#1}}
\newcommand{\eeq}{\end{equation}}

\newcommand{\be}{\begin{equation}}
\newcommand{\ee}{\end{equation}}
\newcommand{\ba}{\begin{array}}
\newcommand{\ea}{\end{array}}

\ifCLASSINFOpdf
\else
\fi
\begin{document}
%

\title{Effect of Batch Normalization on Noise Resistant Property of  Deep Learning Models}
%
%
%
\author{Omobayode~Fagbohungbe,~\IEEEmembership{Student~Member,~IEEE,}
        ~Lijun~Qian,~\IEEEmembership{Senior~Member,~IEEE}
\thanks{The authors are with the CREDIT Center and the Department
of Electrical and Computer Engineering, Prairie View A\&M University, Texas A\&M University System, Prairie View, 
TX 77446, USA. Corresponding author: Lijun Qian, e-mail: liqian@pvamu.edu}}
\maketitle

\begin{abstract}
The fast execution speed and energy efficiency of analog hardware has made them a strong contender for deployment of deep learning model at the edge. However, there are concerns about the presence of analog noise which causes changes to the weight of the models, leading to performance degradation of deep learning model, despite their inherent noise resistant characteristics. The effect of the popular batch normalization layer on the noise resistant ability of deep learning model is investigated in this work. This systematic study has been carried out by first training different models with and without batch normalization layer on CIFAR10 and CIFAR100 dataset. The weights of the resulting models are then injected with analog noise and the performance of the models on the test dataset is obtained and compared. The results show that the presence of batch normalization layer negatively impacts noise resistant property of deep learning model and the impact grows with the increase of the number of batch normalization layers.   
\end{abstract}


%
\IEEEpeerreviewmaketitle

\section{Introduction}
\label{sec:Introduction}

The remarkable success of deep learning (DL) in achieving and surpassing human level performance in cognitive task in recent years has led to its wide adoption in many complex and extremely difficult real-life applications such as computer vision, speech recognition, machine translation, autonomous driving, anomaly detection etc. The unprecedented performance fueling the resurgence of deep learning can be attributed to many factors such as  large scale dataset, high-performance hardware, algorithmic and architectural techniques and more sophisticated optimization methods. These factor have enabled the design of state-of-the-art deep learning models with large and complex architectures.

However, the  fundamental computationally intensive operations performed by deep learning, which increases with model size, has remained the same despite huge improvement in model design~\cite{mixedsignal,Li,joshi,charan}. Also, the data intensive nature of these operations mean that deep learning models also need larger memory and memory bandwidth in order to achieve reasonable performance. Hence, these large models are expensive in terms of computing and memory resources, making them unsuitable for deployment on edge computing devices with limited performance and power budget~\cite{NoisyNN,mixedsignal} such as battery powered mobile and IoT devices. These issues have led to the introduction of specialized energy-efficient computing hardware such as TPU~\cite{tpu} and low power GPUs~\cite{gpu} for deep learning inference. Furthermore, progress has also been made in training  lighter models for this type of hardware using methods such as pruning~\cite{han2015deep,luo2017thinet}, knowledge distillation~\cite{hinton2015distilling}, etc.

Despite these efforts, the maximum achievable energy efficiency has been limited due to CMOS technology approaching its limit. This limit means there's always the presence of the memory wall~\cite{xiao}, a physical separation between the processing unit and the memory, leading to high energy consumption and high latency due to a constant shuttle between the memory and the processing unit for data access~\cite{xiao,joshi}. The large size of deep learning model and required input data means that this bottleneck is significant as the model parameters are stored in the off-chip memory due to the limited capacity of the in-chip memory of the CPU or GPU.

 \begin{figure*}[htbp]
	 \centering
    	 \includegraphics[width=5.8in]{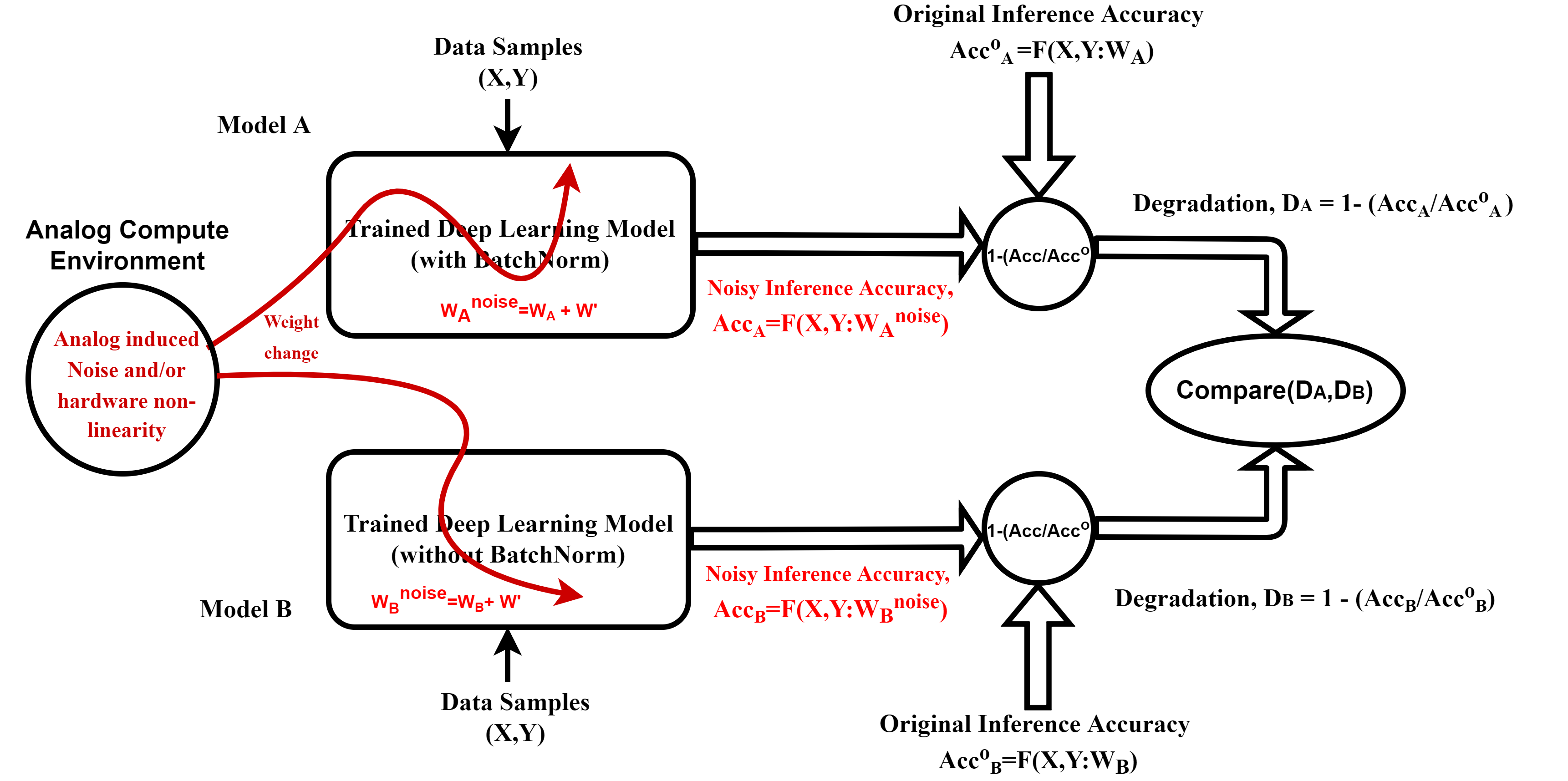}
     	\caption{Comparison of hardware noise induced performance degradation of deep learning models with/without BatchNorm layers. Model A and Model B architecture are the same except Model A has BatchNorm layers, but Model B does not. The better model is the model with lower performance degradation. }
    \label{fig:Hardwareerrorinducedperformancedegradation}
\end{figure*}

In order to achieve better energy efficiency and increase processing speed, there is a growing research interest in specialized, in-memory computing hardware which operates within a tight computational resource and power envelope for model inferencing~\cite{Li}. This specialized hardware must be fast, energy efficient, reliable and accurate~\cite{joshi,mixedsignal}. These tight requirements and advances in analog-grade dense non-volatile memories~\cite{MahmoodiAnalog} are propelling a significant interest in analog specialized hardware for deep learning inference~\cite{NoisyNN}. Analog hardware represents digital values in analog quantities like voltages or light pulses and performs computation in analog domain~\cite{NoisyNN}. This form of computation is cheap and projects a 2X performance improvement over digital hardware in speed and energy efficiency~\cite{Ni,Shen_2017} as they can achieve  projected throughput of multiple tera-operations (TOPs) per seconds and femto-joule energy budgets per multiply-and-accumulate (MAC) operation~\cite{charan,Bennett2020,Burr,marinella}.

\subsection{Challenges and Motivation}
\label{sec:challenges}
Although there are several types, the electronic analog hardware, a form of in-memory computing which encodes network parameters as analog values using non-volatile memory (NVM) crossbar arrays, are the most common. These crossbar arrays enjoy multi-level storage capability and also allow a single time step matrix-vector multiplication to be performed in parallel~\cite{xiao,NoisyNN}, and the addition operation is achieved by simply connecting two wires together where the current add linearly based on Kirchhoff's current law~\cite{NoisyNN,xiao,MITTAL2020101689,joshi}.
However, analog accelerators are imprecise as they do not enjoy bit-exact precision of digital hardware. Furthermore, computation in analog hardware is also very noisy which can lead to degradation in performance of deep learning model or failure during inference~\cite{mixedsignal,NoisyNN}. The factors contributing to computation noise in analog hardware are thermal noise, quantization noise, circuit non-linearity, and device failure~\cite{mixedsignal}. The effect of these factors on the performance of deep learning model was investigated in~\cite{Fagbohungbe2020BenchmarkingIP}. Some of these factors are hard to control and can significantly affect the reliability of DL models or limit their performance despite the known robustness of DL models to analog noise~\cite{Merolla2016DeepNN}. 

Hence, there is a need to understand the effect of current DL model design principles on the noise resistant property of the resulting model. This is expedient in order to design DL models that are more resistant to inherent noise present in analog hardware. Furthermore, it also helps to understand the impact of various design parameters on the model's noise resistant property.

\subsection{Batch Normalization and DL Model Noise Resistant Property}
\label{sec:BatchNorm1}
Batch normalization, or BatchNorm~\cite{Ioffe2015BatchNA}, a popular method used in designing DL model for research and real life application by default, is one of the new techniques enabling the unprecedented performance of DL model in cognitive task. The popularity can be attributed to its ability to ease model training and achieving faster convergence by enabling the use of bigger learning rate, reducing model sensitivity to initialization and also acting as a regularizer~\cite{Ioffe2017BatchRT,Ioffe2015BatchNA,Wu2019L1B}. It achieves these by increasing the network with an extra layer before the activation that aims to stabilize the mini-batch input distribution to a given network layer during training~\cite{Santurkar2018HowDB}. BatchNorm also stabilizes the training process by controlling and setting the  first two moments (mean and variance) of the distribution of each activation to be zero and one, respectively~\cite{Santurkar2018HowDB,Qiao2019MicroBatchTW}. Furthermore, it preserves the model expressivity by normalizing, scaling and shifting the batch input based on the trainable parameters~\cite{Santurkar2018HowDB}. Batch normalization plays an important role in training models that are particularly deep by helping reduce internal covariate shift~\cite{Ioffe2015BatchNA}.

In this work, a systematic study of the effect of batch normalization on the noise resistant characteristics of DL models implemented in analog hardware is carried out as shown in Figure~\ref{fig:Hardwareerrorinducedperformancedegradation}. Specifically, the performance of a trained DL model for image classification task with and without the batch normalization layer in the presence of analog noise is investigated. This is achieved by adding noise of a particular power to the model and measuring its impact on the classification performance. The analog noise in this work is modeled as an additive Gaussian noise which is added to the parameters of the model. 
The contributions of this research work are:
\begin{enumerate}
\item A systematic evaluation of the effect of BatchNorm on the noise resistant property of deep learning model is performed;
\item A theoretical study on why deep learning models have inherent noise resistant property and the contributing factors are provided;
\item A mathematical explanation on how BatchNorm affects the noise resistant property of deep learning models is given;
\item A novel design method for deep learning model with BatchNorm layer is proposed in order to strike a balance between model noise resistant property and model performance.
\end{enumerate}

The remainder of this paper is organized as follows: The theoretical perspective for this work is discussed in Section~\ref{sec:theory}. Section~\ref{sec:methods} contained the various details about the experiment and the results and analysis are presented in Section~\ref{sec:results}. Further discussions and related works are reviewed in Section~\ref{sec:discussion} and Section~\ref{sec:conclusion} concludes the paper.

\section{Theoretical Perspective}
\label{sec:theory}
This section provides a theoretical perspective and mathematical  foundation for this work.

\subsection{Noise Resistant Property of DL Models}
\label{sec:Noise}
There is numerous evidence that attest to the ability of noise to improve the properties or performance of DL models. The effectiveness of noise can be attributed to its ability to help stochastic gradient descent (SGD), the algorithm for training DL models, to converge to good global minima and avoid saddle points or local minima during training. The noise in training takes the form of noise injection into model input, model output, model weights, model pre-activation, and model gradient during training. Hence, the presence/absence of noise during the training process can influence the properties of the resulting DL model. 

The inherent presence of noise in the SGD algorithm used in model training is one of the contributing factors for the noise resistant property of DL models. The presence of inherent noise in the SGD algorithm is established in the derivation below.

Consider a simple model with loss function $L(x)$ defined as the average of the sum of the losses of each sample in the dataset defined in equation (\ref{eequ1}). 
\begin{equation}
\label{eequ1}
L_D(x)=\frac{1}{N}{\sum_{i \in N} {L_i(x)}}
\end{equation}

Since SGD is performed using a fixed batch size which is randomly selected, the equivalent loss over a batch size $B$ is defined as 
\begin{equation}
\label{eequ2}
L_B(x)=\frac{1}{B}{\sum_{i \in B}{L_i(x)}}
\end{equation}

Hence, the gradient of the loss function which uses a learning rate of $\alpha$ is defined mathematically as 
\begin{equation}
\label{eequ3}
\alpha \nabla_{SGD}(x)=\frac{\alpha}{B}{\sum_{i \in B} {\nabla L_i(x)}} 
\end{equation}

By adding and subtracting $\alpha\nabla L_{D}(x)$, equation (\ref{eequ3}) becomes

\begin{equation}
\label{eequ4}
\alpha \nabla_{SGD}(x)= \alpha \nabla L_D(x)  + \frac{\alpha}{B}{\sum_{i \in B} ({\nabla L_i(x) - \nabla L_D(x)})} 
\end{equation}

The first term in equation (\ref{eequ4}) is defined as the gradient and the second term is defined as the error term. The error term is expected to have an unbiased but noisy value due to the uniform batch size. Hence 


\begin{equation}
\label{eequ5}
E[\frac{\alpha}{B}{\sum_{i \in B} ({\nabla L_i(x) - \nabla L_D(x)})}]=0
\end{equation}

The noise quantity, $C$, which is architecture dependent and also  single step estimate of the gradient, is mathematically defined as 
\begin{equation}
\label{eequ6}
C=E[\|{{\nabla L_i(x) - \nabla L_D(x)}}\|^2]
\end{equation}

The upper bound of the noise of the gradient quantity is defined in equation (\ref{eequ7}) by applying  basic algebra and probability theory to equation (\ref{eequ6}). 
\begin{equation}
\label{eequ7}
E[\|{{\alpha\nabla L_B(x) - \alpha\nabla L_{SGD}(x)}}\|^2] < \frac{\alpha^2}{|B|}{C}
\end{equation}

The full proof for mathematical derivation above can be found in \cite{Bjorck2018UnderstandingBN}. 

Equation (\ref{eequ7}) defines the power of the inherent noise in an SGD step which is influenced by the inverse of the batch size, square of the learning rate, and square of the difference between the gradients. The higher the learning rate, the more the power of the noise and invariably better the noise resistant property of the model and vice versa. The noise estimate $C$ also influences the inherent  noise in an SGD step as the learning rate. However, the equation suggests that the batch size of the training process negatively influences the inherent power and hence the noise resistant property of the model. In other words, the bigger the batch size, the worse the noise resistant property of the resulted DL model. The difference in these hyper-parameter values  from one model to another explains why the noise resistant property also differ from one model to another.

In general, the noise resistant property of a deep learning model increases with the power of the noise injected or present during training. However, it is important that the magnitude of the noise be proportional to the original parameters. The model training process will fail to converge if the noise magnitude overwhelms the model parameter values~\cite{NoisyNN,zhou2019toward,wen2018smoothout}.

\begin{table*}
\centering
 \caption{The details of the DL models and dataset used in the experiments.}
\label{table:details}
 \begin{tabular}{|*{6}{c|} }
 \hline
 Model Name & Dataset & Number of Classes & Model Input Dimension & \# Images per class\\  
 \hline\hline
 ResNet\_18 & [CIFAR10,  CIFAR100] & [10,100] & 32*32*3 & [6000,600] \\ 
  \hline
 ResNet\_34 & [CIFAR10,  CIFAR100] & [10,100] & 32*32*3 & [6000,600]\\ 
   \hline
 ResNet\_44 & [CIFAR10,  CIFAR100] & [10,100] & 32*32*3 & [6000,600] \\ 
   \hline
 ResNet\_56 & [CIFAR10,  CIFAR100] & [10,100] & 32*32*3 & [6000,600]\\
    \hline
 VGG\_16 & [CIFAR10,  CIFAR100] & [10,100] & 32*32*3 & [6000,600]\\  
 \hline
\end{tabular}
\end{table*}

\subsection{Batch Normalization and Its Effects on Noise Resistant Property}
\label{sec:BN}
BatchNorm is one of the most widely used and important algorithm in DL that has significantly improve the performance of DL model. It achieves this by enabling the use of higher learning rate and also controlling the first two moments (mean and variance) of the input distribution. This ensures that the value of the activations does not grow uncontrollably, which is possible with increase in model depth. In fact, the experimental result in Table~\ref{????} confirms that models with BatchNorm achieves superior performance when compared with models without batch normalization. 

BatchNorm affects the power of the inherent noise present during the training process and hence, the noise resistant property of the model can be inferred from Equation (\ref{eequ7}). Specifically, BatchNorm lowers the value of $\nabla L_i(x)$, thereby reducing the power of the inherent gradient noise. Although this enables the convergence of the training process, the lower value of the power of the gradient noise means the resulting model has a poor noise resistant property. The absence of the BatchNorm layer means that power of the gradient noise is greater than the case where the BatchNorm is used. The increase in the power of the gradient noise results a model that has a better noise resistant property. It should be noted that it is also possible for the value of the power of the gradient noise grow uncontrollably due to the lack of BatchNorm layer, making convergence impossible. 

In case where BatchNorm is needed, specifically for large datasets and very deep models, 
the influence of the BatchNorm on the model noise resistance is depended on the number of BatchNorm layers in the model. Hence, a version of  a model with BatchNorm layer before every activation function will have poor noise resistant property when compared with another version of the same model with fewer BatchNorm layers. This is because the power of the gradient noise with fewer BatchNorm layer is greater than that of the model with a BatchNorm before every activation function. This is expected as the number of BatchNorm in the model affects the value of  $\nabla L_i(x)$, which in turn determines the power of the gradient noise.  

From the above observations, it is clear that the number of BatchNorm layers is very important as it is a design choice that we could control to strike the balance between model performance property and its noise resistant property. The presence of BatchNorm at every layer means that the model can achieve superior performance comparing to when the BatchNorm is absent. However, this comes at the expense of good noise resistance property. In order to mitigate this effect, some BatchNorm layers can be removed from the model in order to trade reduction in model performance for improvement in noise resistant property. This tradeoff is extremely important because on one hand the model will not train well without the presence of BatchNorm layer, especially when the model is quite deep and the learning task is difficult, on the other hand, two much BatchNorm results in poor noise resistant property. Hence, controlling the tradeoff allows the gradient noise power to grow in a controlled manner to achieve decent model performance and noise resistant property at the same time.

\subsection{Smoothing Effect of Batch Normalization}
\label{sec:smoothing}
The authors in~\cite{Santurkar2018HowDB} stated that the effectiveness of BatchNorm on the training process is due to its ability to reparameterizes the underlying non-convex optimization problem to make its landscape significantly smoother. This reparameterization influences the training process by making the loss function and the gradient of the loss to be more Lipschitz.
A function $f$ is said to be L-Lipschitz if:
\begin{equation}
\label{eequ8}
|{ f(x_1) -  f(x_2)}|] \leq L{\|x_{1}-x_{2}\|}
\end{equation}
for all values of $x_1$ and $x_2$.
These effects, also known as Lipschitz continuity and gradient predictability, makes the gradient descent algorithm to be stable, reliable and predictable. Furthermore, BatchNorm moderates or eliminates the exploding or vanishing gradient phenomena as gradient of models with BatchNorm layer has gradient loss with low range of values in every direction~\cite{Santurkar2018HowDB}. Furthermore, the authors in~\cite{Santurkar2018HowDB} also showed that $l_2$ distance between the two consecutive instances of the loss gradient is significantly lower for model with BatchNorm than that of the models without BatchNorm. The low value of the range and the $l_2$ distance of two consecutive instances of the loss gradient of the models with BatchNorm can be interpreted as the gradient contains noise of low power. However, models without BatchNorm do have loss gradient with very wide range of values, i.e., the gradient contains noise of high power. 
In fact, the relationship between the loss gradient of a model with and without BatchNorm is given in the equation (\ref{eequ9}) below~\cite{Santurkar2018HowDB}.
\begin{equation}
\label{eequ9}
\| \nabla _{y_j}\hat{L}\|^2] \leq \frac{\gamma^2}{\sigma_j^2} \Big(  \| \nabla _{y_j}L\|^2] - \frac{1}{m} {\big \langle{1,\nabla _{y_j}L}\big \rangle}^2 - \frac{1}{m} {\big \langle{\nabla _{y_j}L, \hat{y_j}}\big \rangle}^2  \Big) 
\end{equation}
where $L$ and $\hat{L}$ is the loss of the model with and without BatchNorm, respectively. This shows that the loss gradient of the model with BatchNorm is less than the loss gradient of the model without BatchNorm. Apart from the additive reduction, the large value of $\sigma_j$ for model without BatchNorm means that the loss gradient for model with BatchNorm is significantly lower. Please refer to~\cite{Santurkar2018HowDB} for more explanations and derivation of equation (\ref{eequ9}).

This significant difference in the properties of the model with and without BatchNorm layer explains the difference in the noise resistant property of the models. The presence of noise in the loss gradient demonstrates that the model performance and noise resistant properties are influenced by the noise.  This explanation is in agreement with the explanations in Section~\ref{sec:BN} although the present of the noise is accounted for in another manner.

\section{Experimental Setup}
\label{sec:methods}
The discussion in this section focuses on the details of the experiments carried out to study the effect of batch normalization layer on the inherent resistance characteristics of deep learning models to analog noise. The details discussed are the dataset, model design and architecture, methodology, and software and hardware used.

\subsection{Dataset}
\label{subsec:ExpSetup}
The CIFAR10 and CIFAR100 datasets are used for the classification task in this paper and the details about the dataset are provided in Table~\ref{table:details}. CIFAR10 and CIFAR100 datasets are  labelled datasets that are part of  80 million tiny images datasets. Furthermore, each datasets contains 60,000 images divided into 50,000 training and 10, 000 testing datasets.The images in the dataset are of dimension $32\times32\times3$. The images in CIFAR10 dataset can be grouped into 10 classes, with each class containing 5000 training images and 1000 testing images. The classes are mutually exclusive as there is no semantic overlap between the classes.  The CIFAR100 dataset contains images that can be grouped into 100 classes with each class containing 500 training images and 100 testing images. The classes in the CIFAR100 dataset are not mutually exclusive and it can be said that there is some form of semantic overlap as the dataset contain 20 superclass.  

\subsection{Model Design and Training}
\label{subsec:ExpSetup}
The models used for this work are stated in Table~\ref{table:details}. The models are all models suitable for image classification task as CIFAR10 and CIFAR100 datasets can only be used for such task. The models can generally grouped into ResNet and VGG16 models. ResNet models are form of convolutional neural network based on the residual learning framework which eases the training and optimization of deeper models. It achieves this by reformulating the constituent layers of a model as learning residual function with reference to the layers inputs instead of learning unreferenced function~\cite{resnet}. The VGG16 is also a form of CNN that leverages on the convolutional network depth  using an architecture with very small (3x3) convolution filters to achieve better performance~\cite{vgg}.
All the models used in this work were trained from scratch until convergence, as the model weights were not initialized using weights from other models. With glorot-uniform method as the initializer, categorical cross entropy as the loss function and Adams optimization algorithm as the optimizer, model convergence was achieved by minimizing the prediction error and maximize the model accuracy. Data augmentation was also performed to prevent overfitting and maximize model performance.

\subsection{Model Training Stage}
\label{sec:training}
A deep learning model with  batch normalized layers is trained using the selected dataset from scratch. The deep learning model architecture is designed such that there is one batch normalization layer after every convolutional and fully connected layer in the model. The model is trained until convergence is achieved and this model is considered as the baseline model. All the models under consideration with BatchNorm layer achieve model convergence.
Afterwards, the same deep learning model without the batch normalization is trained from scratch until convergence is achieved. This is the experience with all the ResNet models without the BatchNorm layer as they achieve convergence when trained with CIFAR10 dataset.

If the model convergence is not achieved, the model architecture is modified slightly by inserting a minimum number of batch normalization layers required for convergence to be achieved. This is the case referred as partial presence of BatchNorm layer in this work. The minimum number of BatchNorm layer required varies from one model to another and also from one dataset to another, hence a trial and error method is used to determine it. A choice of trial and error method to determine this minimum number of batch normalization layer(s) need to achieve convergence for a particular model is informed by the non-trivial nature of the problem. In this work, this approach is used to achieve convergence for all the ResNet model without BatchNorm layer when trained on CIFAR100 dataset and VGG16 model when trained on both CIFAR10 and CIFAR100 dataset.

\subsection{Model Inference stage}
\label{sec:inference}
The inference performance of the models obtained from section~\ref{sec:training} in the presence of analog noise is investigated in this section.  The performance metric for this work is the classification accuracy (\%), a common metric for deep learning models for image classification task. The analog noise is modeled as additive white Gaussian noise which is added to the model weights. The weights of the model due to the presence of noise is represented in the equation (\ref{equ1}) below.

The white Gaussian noise used in this work is of zero mean and a standard deviation of $\sigma_{noise}$, which represents the energy of the noise. The value of $\sigma_{noise}$ is calculated using equation (\ref{equ1}) below where $SNR$ is the signal to noise ratio and $\sigma_w$ is the standard deviation of the weights in a layer.
\begin{equation}
\label{equ1}
{\sigma_{noise}}=\frac{\sigma_w}{SNR}
\end{equation}
Equation (\ref{equ3}) is obtained by substituting equation (\ref{equ2}) into equation (\ref{equ1}) where $\eta$ is defined as a noise form factor. 
\begin{equation}
\label{equ2}
{\eta}=\frac{1}{SNR}
\end{equation}
\begin{equation}
\label{equ3}
{\sigma_{noise}}={\eta}\times{\sigma_w}
\end{equation}
The $SNR$ values of the standard deviation of the Gaussian noise used in this work are 100, 10, 5, and 2.5. These $SNR$ values are equivalent to Gaussian noise of zero mean and standard deviations equivalent to 1\%, 10\%, 20\%, and 40\% form factor of the standard deviation of the weights of a particular layer $\sigma_w$.
The models with batch normalization obtained in Section \ref{sec:training} above is put in inference mode and the performance of the model on the test dataset is evaluated in order to obtain the model classification accuracy. The classification accuracy obtained for the model without the analog noise ($\eta$=0) is the baseline inference accuracy for the model. The additive Gaussian noise of zero mean and the desired standard deviation  equal to 1\% of the $\sigma_w$ at a layer 1 is added to weights in layer 1 of the model. This is equivalent to $SNR$ value of 100\% and $\eta$ value of 1\%. This procedure is repeated for all the layers in the model until noise is added to all the weights in the model. The performance of the model with the new weight is then evaluated using the test dataset to obtain the inference classification due to the noise. The procedure above for a fixed value of $\eta$ is then repeated multiple times and the average inference classification accuracy recorded. The average classification accuracy due to the present of the noise is then normalized with the baseline inference classification accuracy using the formulae in equation (\ref{equ4}).
 \begin{equation}
 \label{equ4}
A_1=\frac{a_1}{a_o}
\end{equation}
where $a_o$ are the baseline classification accuracy, $A_1$ and $a_1$ are the normalized classification accuracy and average classification accuracy due to the present of noise of $\eta$ value of $1$. These procedures are then repeated for $\eta$ values of 10\%, 20\%, 30\% and 40\% and the corresponding average and normalized classification accuracy noted. The procedure above is then repeated with the model without batch normalization layer for the same $\eta$ values above and the corresponding classification accuracy recorded.    

\subsection{Software and Hardware}
\label{subsec:ExpSetup}
Keras deep learning framework, using tensorflow backend  is used for training and testing all the models in this work. The Keras framework was installed on the NVIDIA V100-DGXS-32GB GPU.

\section{RESULTS and ANALYSIS}
\label{sec:results}
\begin{table*}
\centering
\caption{Comparison of the performance of VGG16 and ResNet Models with and without Batch normalization layer in the presence of noise in all its layer during inference when tested with CIFAR10 dataset. The performance metric is the model classification accuracy.}
\label{tab:result_cifar10}
\begin{tabular}{|c|c|c|c|c|c|c|c|c|c|c|c|c|} 
\hline
\multirow{3}{4em}{} & \multicolumn{12}{|c|}{Noise factor, $\eta$}\\ \hline

& \multicolumn{2}{|c|}{0\%}& \multicolumn{2}{|c|}{1\%}& \multicolumn{2}{|c|}{10\%}& \multicolumn{2}{|c|}{20\%}& \multicolumn{2}{|c|}{30\%}& \multicolumn{2}{|c|}{40\%}\\ \hline

{Model Name} & {With BN} & {No BN}& {With BN}& {No BN}& {With BN} & {No BN}& {With BN}& {No BN}& {With BN} & {No BN}& {With BN} & {No BN}\\ \hline

{Resnet\_20} & \textcolor{blue}{92.16\%} & {89.62\%}& \textcolor{blue}{91.48\%}& {89.60\%}& \textcolor{blue}{\textbf{87.20\%}}& \textbf{88.49\%}& \textcolor{blue}{66.16\%}& {84.90\%} &\textcolor{blue} {26.47\%}& {76.65\%} &\textcolor{blue}{13.83\%}& {61.37\%}\\ \hline

{Resnet\_36} & \textcolor{blue}{92.46\%} & {89.25\%}& \textcolor{blue}{90.46\%}& {89.27\%}& \textcolor{blue}{\textbf{81.65\%}} & \textbf{88.13\%}&\textcolor{blue} {47.18\%}& {84.30\%}& \textcolor{blue}{20.94\%}& {75.62\%}& \textcolor{blue}{12.59\%} & {62.67\%}\\ \hline

{Resnet\_44} & \textcolor{blue}{91.81\%} & {87.99\%}& \textcolor{blue}{85.00\%}& {88.06\%} & \textcolor{blue}{\textbf{57.67\%}}& \textbf{87.32\%}& \textcolor{blue}{22.55\%}& {84.11\%} & \textcolor{blue}{14.74\%}& {77.71\%} & \textcolor{blue}{9.73\%}& {67.32\%}\\ \hline

{Resnet\_56} & \textcolor{blue}{92.71\%} & {87.78\%}& \textcolor{blue}{88.40\%}& {87.81\%}& \textcolor{blue}{\textbf{31.81\%}} & \textbf{86.74\%}& \textcolor{blue}{22.02\%}& {82.85\%}& \textcolor{blue}{12.67\%} & {74.41\%}& \textcolor{blue}{11.09\%} & {61.77\%}\\ \hline

{VGG\_16} & \textcolor{blue}{93.06\%} & {93.02\%}& \textcolor{blue}{10.00\%}& {93.01\%}& \textcolor{blue}{\textbf{10.00\%}} & \textbf{92.30\%}& \textcolor{blue}{10.00\%}& {89.84\%}& \textcolor{blue}{10.00\%} & {89.84\%}& \textcolor{blue}{10.00\%} & {76.17\%}\\ \hline
\end{tabular}
\end{table*} 

\begin{figure*}[htbp]
	 \centering
    	 \includegraphics[width=18cm]{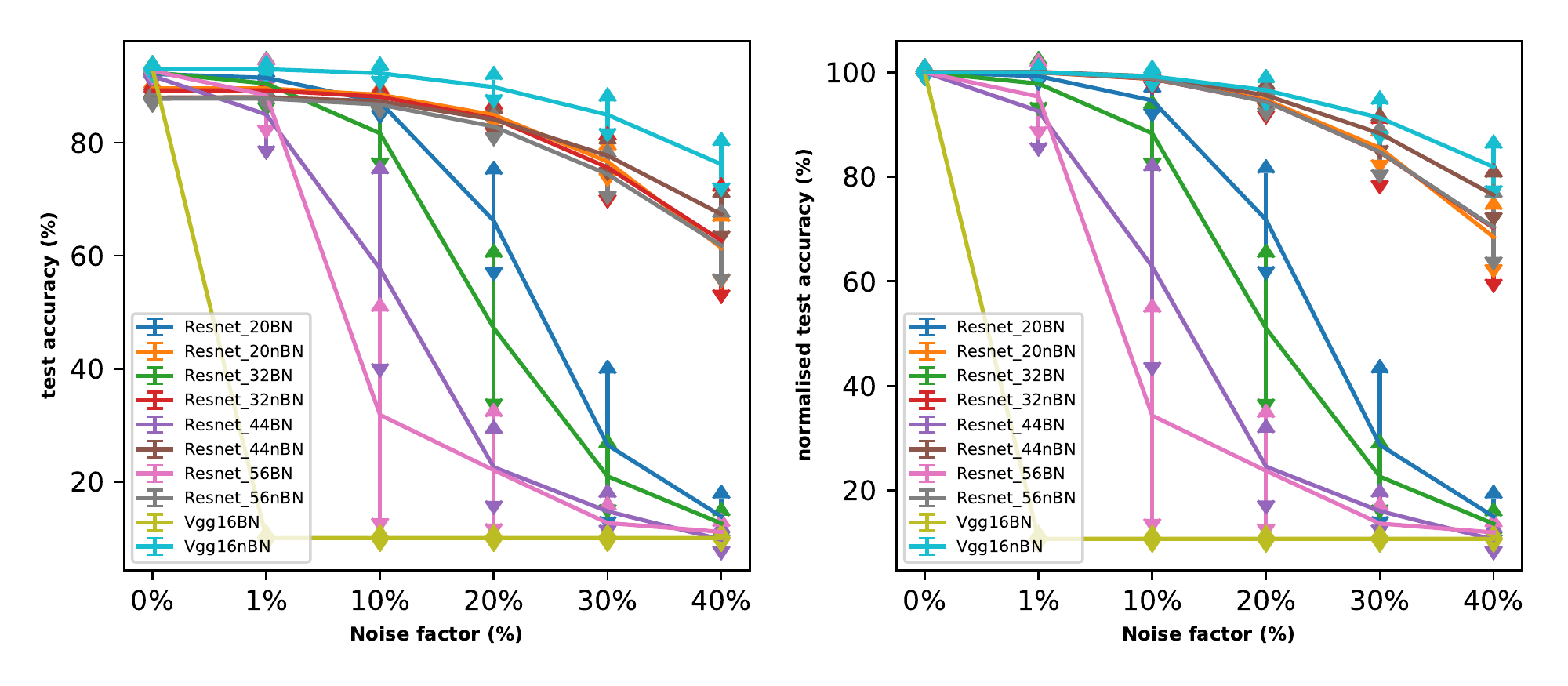}
     	\caption{The inference performance comparison of pre-trained VGG16 model and ResNet models trained with and without batch normalization layer trained on CIFAR10 dataset when Gaussian noise is added to all the weights. The performance metric is the (a) Actual average test accuracy(b)Normalised average test accuracy}
    \label{cifar10NormalisedImage}
\end{figure*}

The results and analysis of the experimental work to understand the effect of batch normalization on the robustness of deep neural network to analog noise using CIFAR10 and CIFAR100 dataset are presented in this section. This effect is studied by comparing the performance of a deep learning model with batch normalization layer after each layer (convolutional and fully connected layer) and the same model architecture without the batch normalization layer after each layer or partially present in some layers. The partial presence of batch normalization is a situation where batch normalization layer are present in some layers in order to kick-start the training process. This generally apply to situation where models with reasonable performance are difficult to train without batch normalization due to classification task complexity and/or model architecture. In this work, the use of partial presence of batch normalization is only used as substitute for the full absence of batch normalization for the VGG16 model training on CIFAR10 dataset and all models on the CIFAR100 dataset. It should be noted that the minimum number of batch normalization layers needed to train the models varies from one use case to another.

\begin{table*}
\centering
\caption{Comparison of the performance of VGG16 and ResNet Models with and without Batch normalization layer in the presence of noise in all its layer during inference when tested with CIFAR100 dataset. The performance metric is the model classification accuracy.}
\label{tab:result_cifar100}
\begin{tabular}{|c|c|c|c|c|c|c|c|c|c|c|c|c|} 
\hline
\multirow{3}{4em}{} & \multicolumn{12}{|c|}{Noise factor, $\eta$}\\ \hline

& \multicolumn{2}{|c|}{0\%}& \multicolumn{2}{|c|}{1\%}& \multicolumn{2}{|c|}{10\%}& \multicolumn{2}{|c|}{20\%}& \multicolumn{2}{|c|}{30\%}& \multicolumn{2}{|c|}{40\%}\\ \hline

{Model Name} & {With BN} & {No BN}& {With BN}& {No BN}& {With BN} & {No BN}& {With BN}& {No BN}& {With BN} & {No BN}& {With BN} & {No BN}\\ \hline

{Resnet\_20} &  \textcolor{blue}{66.69\%} & {63.07\%}&  \textcolor{blue}{\textbf{66.61\%}}& \textbf{62.90\%}&  \textcolor{blue}{55.63\%} & {54.84\%}&  \textcolor{blue}{26.38\%}& {34.98\%}&  \textcolor{blue}{8.43\%} & {16.12\%}&  \textcolor{blue}{2.69\%} & {5.80\%}\\ \hline

{Resnet\_36} &  \textcolor{blue}{69.60\%} & {66.84\%}&  \textcolor{blue}{\textbf{69.00\%}} & \textbf{66.69\%}&  \textcolor{blue}{43.42\%} & {57.42\%}&  \textcolor{blue}{21.97\%}& {27.25\%}&  \textcolor{blue}{6.78\%} & {7.44\%}&  \textcolor{blue}{1.78\%} & {2.66\%}\\ \hline

{Resnet\_44} &  \textcolor{blue}{69.50\%} & {65.47\%}&  \textcolor{blue}{\textbf{68.45\%}}& \textbf{65.42\%}&  \textcolor{blue}{34.85\%} & {58.95\%}&  \textcolor{blue}{12.34\%}& {36.76\%}&  \textcolor{blue}{4.01\%} & {11.20\%}& \textcolor{blue} {1.80\%} & {2.86\%}\\ \hline

{Resnet\_56} &  \textcolor{blue}{69.54\%} & {66.53\%}&  \textcolor{blue}{\textbf{68.27\%}}& \textbf{66.20\%}&  \textcolor{blue}{3.29\%} & {53.50\%}&  \textcolor{blue}{1.64\%}& {23.40\%}&  \textcolor{blue}{1.25\%} & {3.94\%}&  \textcolor{blue}{1.16\%}& {1.85\%}\\ \hline

{VGG\_16} &  \textcolor{blue}{69.65\%} & {66.02\%}&  \textcolor{blue}{\textbf{1.00\%}}& \textbf{66.07\%}&  \textcolor{blue}{1.00\%} & {1.00\%}&  \textcolor{blue}{1.00\%}& {1.00\%}&  \textcolor{blue}{1.00\%} & {1.00\%}&  \textcolor{blue}{1.00\%} & {1.00\%}\\

 \hline
\end{tabular}
\end{table*}

\begin{figure*}[htbp]
	 \centering
    	 \includegraphics[width=18cm]{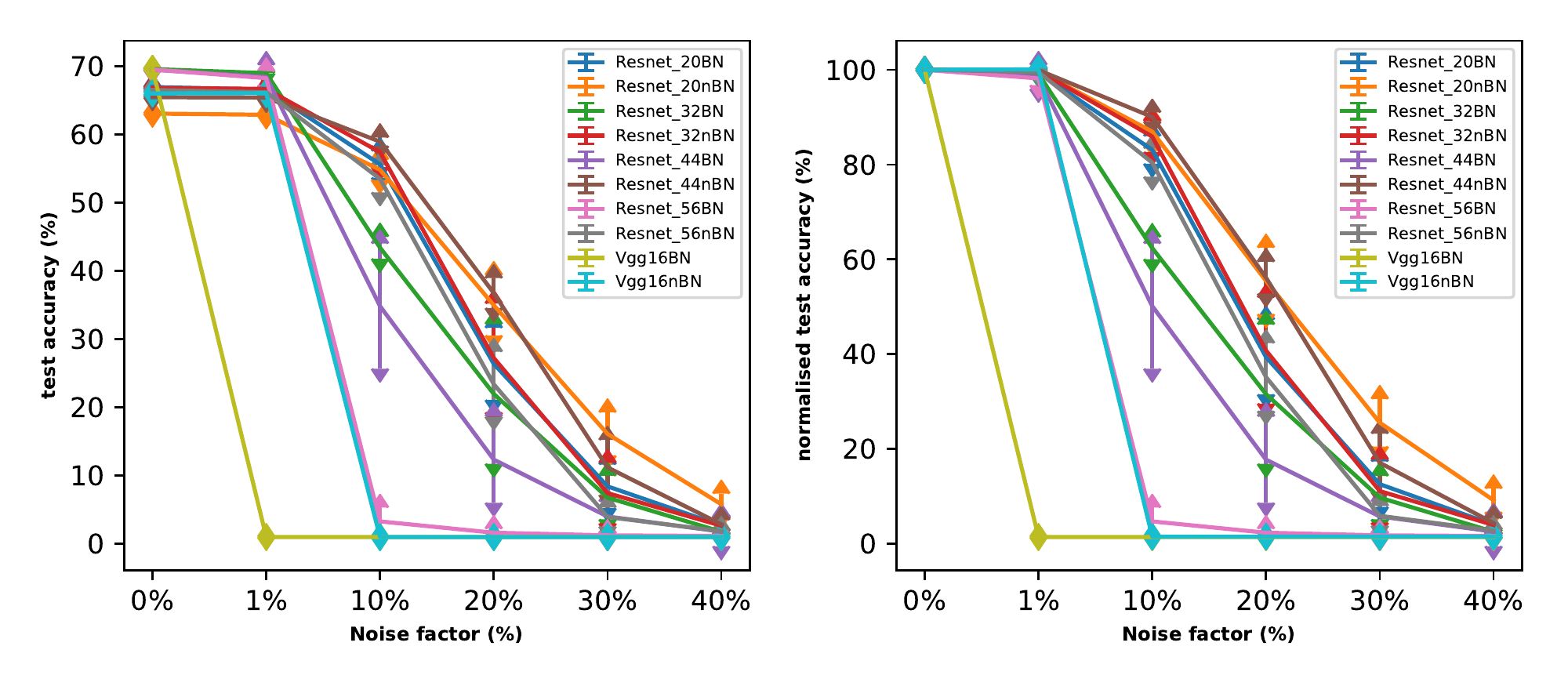}
     	\caption{The inference performance comparison of pre-trained VGG16 model and ResNet models trained with and without batch normalization layer trained on CIFAR100 dataset when Gaussian noise is added to all the weights. The performance metric is the (a) Actual average test accuracy(b)Normalized average test accuracy}
    \label{cifar100NormalisedImage}
\end{figure*}

\subsubsection{Results with CIFAR10}
\label{sec:cifar10_result}
The results of the experimental work is presented in Tables \ref{tab:result_cifar10} and Figures \ref{cifar10NormalisedImage}, and it shows the testing accuracy and normalized testing accuracy when analog noise is added to all the weights of the model respectively. Firstly, the baseline testing accuracy of the models with BatchNorm, which is equivalent to analog noise with noise factor ($\eta$) of 0\%, is slightly higher than baseline testing accuracy of the models without BatchNorm. This is expected as batch normalization eases model training and achieves faster convergence by enabling the use of bigger learning rate, reduce model sensitivity to initialization and also acts a regularizer~\cite{Ioffe2017BatchRT, Ioffe2015BatchNA,Wu2019L1B}. 

Secondly, it is observed that model inference performance of the models (with or without BatchNorm) decreases with increase in the analog noise power level. This trend is in agreement with the observation of the authors in~\cite{Fagbohungbe2020BenchmarkingIP} as the ability of the noise to corrupt the model weight increases with the noise power level. For example, ResNet56 with the BatchNorm has an inference accuracy value of 
92.17\%, 88.40\%, 31.81\%, 22.02\%, 12.67\%, and 11.09\% at noise power level of 0\%, 1\%,10\%, 20\%, 30\% and 40\% respectively.  This trend is also the same with ResNet56 without the BatchNorm layer where an inference accuracy of 87.88\%, 87.81\%, 86.74\%, 82.85\%, 74.4\%, and 61.77\% at noise power level of 0\%, 1\%, 10\%, 20\%, 30\% and 40\% respectively.

However, the severity of the effect of the noise power level on the model inference accuracy is actually more for models with BatchNorm layer than models without BatchNorm layer. This severity is observed when comparing the normalized inference accuracy for the models at various noise level, which is shown in Figures \ref{cifar10NormalisedImage}b. The normalized inference accuracy of the model is obtained by normalizing average inference accuracy of a model by the baseline inference accuracy of the model as stated in equation (\ref{equ4}). The high percentage performance degradation suffered by models with batch normalization layer in the presence of analog noise means that the model loses their initial performance advantage in the form of better inference classification accuracy. In fact, the classification accuracy of the model with batch normalization layer reduces significantly below the classification accuracy of the model without the batch normalization layer particularly at higher noise factor. This trend can be observed in Figure \ref{cifar10NormalisedImage} a where the inference accuracy for most of models with the batch normalization layer has degraded below the inference accuracy of the models without the batch normalization layer by the noise factor of 10\%. At noise power level of 10\%, the normalized inference accuracy of ResNet56 model with BatchNorm and ResNet56 model without BatchNorm layer is 34.31\% and 98.81\% respectively. This behavior is also observed with VGG16 model where the model with BatchNorm and Model without BatchNorm has a normalized inference accuracy of 10.75\% and 91.31\% respectively. This shows that models with BatchNorm layer has poor noise resistant property as compared with the models without BatchNorm layer.  

Furthermore, It is also observed that shallower ResNet models with BatchNorm are more noise resistant than deeper models ResNet model with BatchNorm layer. At noise power level of 10\%, ResNet20 has the best normalized inference accuracy of 94.62\% while ResNet56 has the poorest normalized inference accuracy of 34.31\%. This behavior is not observed in ResNet models with BatchNorm layer where the normalized inference accuracy is fairly constant for 10\% noise power level. This is behavior might be due to reduced influence of BatchNorm in those models. This reduced influence is due to the reduced number of BatchNorm layer present in the model architecture as compared to deeper models.

\subsubsection{Results with CIFAR100}
\label{sec:cifar100_result}
Table \ref{tab:result_cifar100} and Figure \ref{cifar100NormalisedImage} shows the inference accuracy and normalized inference accuracy of the various models pre-trained on CIFAR100 in the presence of analog noise of different noise power level. Like the case with CIFAR10, the models with BatchNorm have a BatchNorm layer after each convolution and fully connected layer. However, the all the models without BatchNorm has a few layers with BatchNorm layer due to the inability of the model to achieve model convergence without it. The number is however limited to the minimum required to kickstart the training process and this minimum is obtained by trial and error. 

The CIFAR100 baseline inference accuracy of all the models is lower than baseline for CIFAR100 dataset. The is because the classification task on CIFAR100 data is more complex than that of CIFAR10 dataset as the CIFAR100 contain 100 classes as compared to 10 classes in CIFAR10 dataset. The semantic overlap between the classes in CIFAR100, unlike CIFAR10 where there is none, also makes the classification task even non-trivial. Furthermore, the number of images per class is smaller for CIFAR100 (600) than CIFAR10 (6000) making the task even more difficult.

The baseline inference accuracy (noise power level of zero) of the models with the BatchNorm layer is significantly higher than "No BatchNorm" counterparts despite having some BatchNorm layer themselves. This difference is behavior is due to the limited influence of BatchNorm on the "No BatchNorm" models as these models have very few BatchNorm layers as compared with models with BatchNorm layer which has a BatchNorm layer after every convolution and fully connected layer.  

However, models with BatchNorm layer suffers more degradation in performance in the presence of additive analog noise compared to models "without BatchNorm"  and this degradation performance also increases with increase in the noise power level. This degradation is showed in Figure \ref{cifar100NormalisedImage}b where the normalized inference accuracy of the models with BatchNorm are lower than the normalized inference accuracy of models "without BatchNorm". For all ResNet models with BatchNorm, shallower models have better noise resistant property than deeper models as indicated by higher normalized noise resistant property for a fixed noise power level due to them having fewer number of BatchNorm layer. These observations is also observed with these models trained on CIFAR10 dataset and detailed in section \ref{sec:cifar10_result}. The differences lie in the severity of the degradation experienced by models trained on CIFAR100 dataset which is discussed later.

The significant difference in the baseline inference accuracy of models with BatchNorm and models "without BatchNorm"  trained on CIFAR100 dataset means that the absolute accuracy of the model with BatchNorm layer might be higher than model without BatchNorm layer at some noise power level despite the former having poorer noise resistant property. This is true for all ResNet models without BatchNorm trained on CIFAR100 dataset at noise power level of 1\% and lower. This means these models with BatchNorm can be selected ahead of their "No BatchNorm" counterpart to be implemented on analog computing device. This also applies to models ResNet20 and ResNet36 models with BatchNorm trained on CIFAR10 dataset as their absolute inference accuracy is still higher than inference value of their No BatchNorm variant despite their performance degradation in the presence of noise.

\subsubsection{Impact of Model Task on its Noise Resistant Property}
\label{sec:cifar10_cifar100_result}

\begin{figure*}[htbp]
	 \centering
    	 \includegraphics[width=18cm]{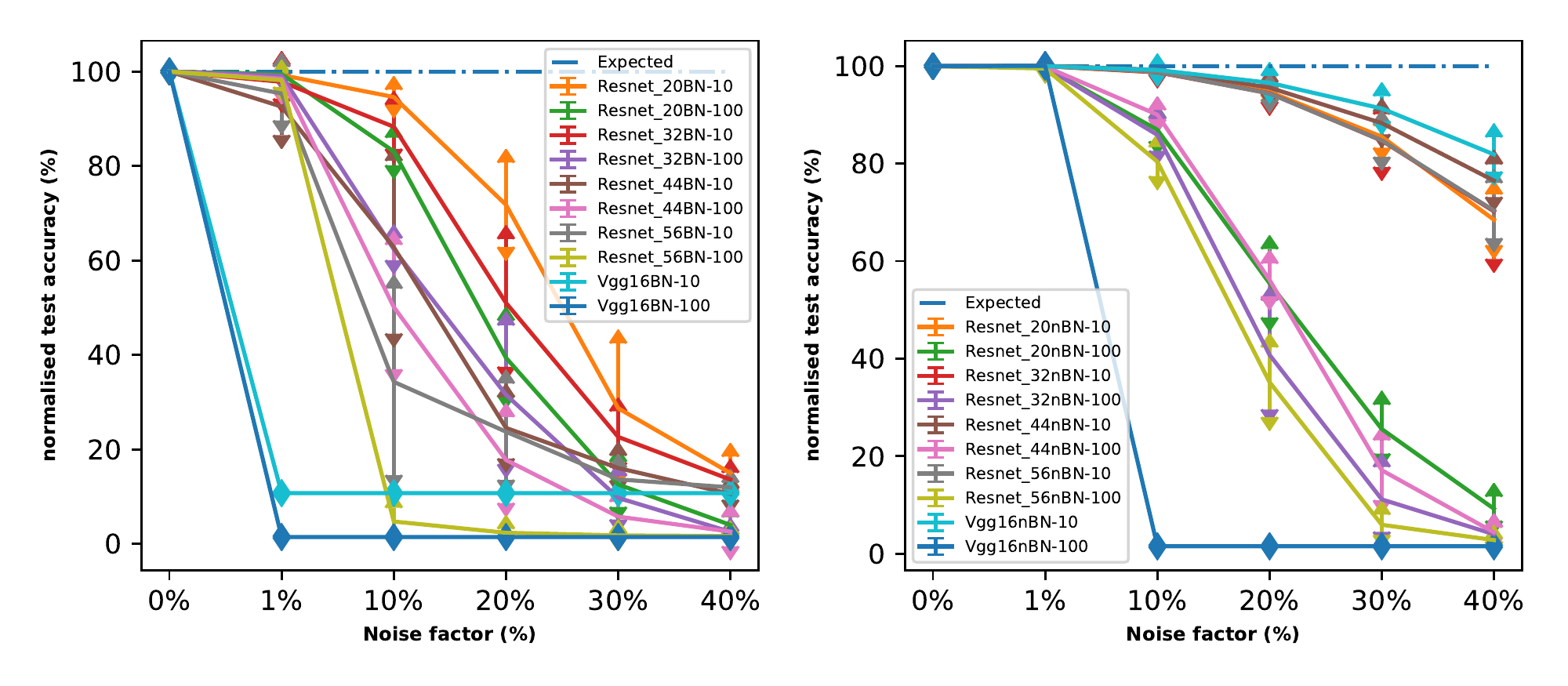}
     	\caption{(a)The inference performance comparison of pre-trained VGG16 model and ResNet models trained with batch normalization layer on both CIFAR10 and CIFAR100 dataset when Gaussian noise is added to all the weights. The performance metric is the actual average test accuracy(b)The inference performance comparison of pre-trained VGG16 model and ResNet models without batch normalization layer trained on both CIFAR10 and CIFAR100 dataset when Gaussian noise is added to all the weights. The performance metric is the normalized average test accuracy}
    \label{cifar10-cifar100-normalised}
\end{figure*}

The impact of the complexity of the model's task on the noise resistant property of the models are analyzed in this section. A performance comparison between the models trained on the CIFAR10 and CIFAR100 datasets using their normalized inference accuracy when Gaussian noise is injected into their model weight is shown in Figure \ref{cifar10-cifar100-normalised}. While the comparison is done with models with batch normalization layer trained on CIFAR10 and CIFAR100 datasets in \ref{cifar10-cifar100-normalised}a, the comparison in Figure \ref{cifar10-cifar100-normalised}b is between models without or partial batch normalization layer. It is observed from Figure \ref{cifar10-cifar100-normalised}a that models trained on CIFAR10 dataset suffered less percentage performance degradation as compared to models trained on CIFAR100 dataset. This performance difference can be attributed to the complexity of the task as both models contain the same amount of batch normalization layer. This trend is also noticed in Figure \ref{cifar10-cifar100-normalised}b except that models In Figure \ref{cifar10-cifar100-normalised}a suffers more degradation due to the presence of BatchNorm layer after every  in their model architecture which negatively impacts model noise resistant property. Although some models in Figure \ref{cifar10-cifar100-normalised} contain some BatchNorm layer, the number is few, thus limiting their impact on the noise resistant property of the models.

\subsubsection{Trade off between Performance and  Noise Resistant Property}
\label{sec:tradeoff}
It can be observed that the testing accuracy of the model without batch normalization layer is lower than the model with batch normalization layer for both CIFAR10 and CIFAR100 datasets a expected. However, it is possible to get better results by exploring the partial batch normalization case in order to achieve better result by gradually increasing the number of batch normalization layer in order to achieve higher baseline accuracy. However, these improved in the baseline inference accuracy is achieved at the expense of the robustness of the resulting model to Gaussian  noise. Hence, the current practice of using a batch normalization layer at the top of every convolutional might not be the best for models that needs to be deployed on analog hardware as the robustness of the resulting model might be low. The use of partial batch normalization might be better as we can gradually find the number of batch normalization layer needed to achieve a desired performance value. This design philosophy helps us to find some tradeoff between performance and robustness to noise.

In order to aid better summarization and analysis of results, a new metric called average normalized percentage classification accuracy is introduced. This is mathematically defined as:
\begin{equation}
A_{avr}=\frac{\sum\limits_{i=1}^{N} A_i}{N}
\end{equation}
where $A_{avr}$ is the average normalized percentage classification accuracy and $N$ is the number of non-baseline noise factor which is 5. This new metric summarizes the performance of all the model trained on the CIFAR10 and CIFAR100 datasets over the 5 non-baseline noise factor and the result is given in Table \ref{tab:result_cifar100-cifar10}.VGG16 model has the best performance among all the models trained on CIFAR10 dataset and ResNet\_20 model has the best performance among all the models trained on CIFAR100 dataset as they suffer a performance degradation of 6.20\% and 54.62\% respectively.

\begin{table}
\centering
\caption{Comparison of the average normalized percentage classification of VGG16 and ResNet Models with and without Batch normalization accuracy over 5 non-baseline noise factor in the presence of Gaussian noise in all its layer during inference when tested with CIFAR10 and CIFAR100 dataset. The performance metric is the model classification accuracy.}
\label{tab:result_cifar100-cifar10}
\begin{tabular}{|c|c|c|c|c|c|c|c|c|c|c|c|c|} 
\hline
\multirow{3}{4em}{} & \multicolumn{4}{|c|}{Dataset}\\ \hline

& \multicolumn{2}{|c|}{CIFAR10}& \multicolumn{2}{|c|}{CIFAR100}\\ \hline

{Model Name} & {With BN} & {No BN}& {With BN}& {No BN}\\ \hline

{Resnet\_20} & {61.88\%} & {89.49\%}& {47.71\%}& {55.38\%}\\ \hline

{Resnet\_36} & {54.69\%} & {89.63\%}& {41.08\%}& {48.31\%}\\ \hline

{Resnet\_44} & {41.32\%} & {91.95\%}& {34.95\%}& {53.52\%}\\ \hline

{Resnet\_56} & {35.81\%} & {89.68\%}& {21.74\%}& {44.76\%}\\ \hline

{VGG\_16} & {10.75\%} & {93.80\%}& {1.44\%}& {21.23\%}\\

 \hline
\end{tabular}
\end{table}


\section{Related Works and Discussions}
\label{sec:discussion}

Batch normalization was introduced in~\cite{Ioffe2015BatchNA} to dramatically accelerate the training and improve model performance by reducing the internal covariate shift in layers or subnetwork of a deep learning model. It achieves this by performing normalization for each training mini-batch, allowing the use of higher learning rate and making the process insensitive to model initialization. However, batch normalization performance reduces when training batch is small or do not consist of independent samples. Furthermore, the low bit-width quantization technique impedes fundamental mathematical operations in batch normalization in addition to needing additional computation and memory resource. Many methods have been proposed to resolve these issues including batch renormalization~\cite{Ioffe2017BatchRT}, group normalization~\cite{Wu2018GroupN}, layer normalization~\cite{Ba2016LayerN} , weight normalization~\cite{Salimans2016WeightNA}, L1-Norm batch normalization~\cite{Wu2018GroupN}, etc.

The noise resistant ability of neural networks has attracted significant attention recently due to the needs to deploy deep learning models on analog accelerators that contain significant noise. There are extensive works in the literature that relate to analog noise and neural network models. The use of noisy input to improve the ability of neural network to generalize to previous unseen data, recognize faulty input and improve their fault tolerant ability has been discussed  in~\cite{SIETSMA199167,Minnix,Meng, Rusiecki2014TrainingNN}. Deep noise injection, injecting noise into the model weight during training, to improve the noise resistant ability of neural network have been proposed in~\cite{Murray, Qin,Qin2018TrainingRN,Miyashita}. The use of Langevin noise which is adding noise to the weight change to improve model generalization is discussed in~\cite{An}. Furthermore, the work by~\cite{Bishop,Matsuoka,An,Holmstrom,noh2017regularizing} also provides the various theoretical and mathematical justification for use of analog noise for training neural network for the various scenarios mentioned earlier. 

(Check this paragraph for correctness:) Recently, knowledge distillation, a training method which involves learning one model  from another model, and deep noise injection is further used to improve on the existing result \cite{NoisyNN}. Also, method to improve the robustness of neural network against adversarial noise was studied in~\cite{Zheng2016ImprovingTR}  and the generalization of neural network trained with with noisy labels was discussed in~\cite{Chen2019UnderstandingAU,Reed2015TrainingDN}.

The robustness of neural network models to noise can also be achieved by exposing the model to circuit nonlinearities and other constraints during training as done in~\cite{Bo,schmid}. By using the same analog  hardware during training and testing, the model is being conditioned to give good performance in a noisy computational environment. Despite the effectiveness of this method, implementing low power training procedure on analog device is non-trivial and too laborious for a procedure needed only once. Furthermore, implementing the circuitry needed for backpropagation leads to increase in chip's area and complicates chip design~\cite{mixedsignal}. Chip-in-the-loop method, a method suited for inference-only hardware, is introduced in~\cite{Bayraktaroglu,NeuroSmith} by adapting pre-trained model weights for the inference only chip. This process is slow and inefficient and it involves programming the pre-trained weight of a model to the hardware of interest in order to measure the precise error using forward computation pass. The measured error is then used to update the weights of the pre-trained model via backpropagation in software using traditional processors~\cite{mixedsignal}. The work in~\cite{Garg2021DynamicPA} also extend the analog computing architectures to support dynamic precision with redundant coding by repeating operations and averaging the result.

The protection of the weights and bias of neural network from noise using linear and nonlinear analog error correction codes in order to prevent performance degradation has been proposed in~\cite{Upadhyaya2019ErrorCF,Upadhyaya2019ErrorCF2}. The works also explored the use of unequal error protection method for weights at different layers of a binarized network due to the uneven effect of noise in different layers. Furthermore, alternative binary representation of the parameters and weight nulling, a simple parameter error detection method, is proposed in~\cite{QinBitFlipping} to mitigate the effect of the distortion caused by bit flips to improve on model robustness. An algorithm based on deep reinforcement learning which uses selection protection scheme to choose critical bit for error correction code (ECC) protection is developed in~\cite{huang2020functional}.  This critical bit, which is not always the most significant bit, is selected in order to achieve optimal tradeoff between ECC'S redundancy and neural network model performance. The algorithm is a form of function oriented ECC algorithm, achieves this by using an optimization function that optimizes the neural network performance after error correction instead of minimizing the uncorrectable bit error rate in the protected bits. The work in~\cite{LiuDAC2019} uses a collaborative logistic classifier which leverages asymmetric binary classification coupled with an optimized variable-length decode-free  Error-Correcting Output Code to improve the error-correction capability of DNN accelerators. The work in~\cite{Zhang2019} introduces the use of a generalized fault aware pruning technique to improve model resilience to hardware fault. Lastly, the work in~\cite{ShieldeNN} introduces a framework that synergies the mitigation approach and computational resources using neural network architecture-aware partial replacement approach to identify the parameters of interest in the consecutive network layers for hardening to improve model resilience.

The existing works are different from this work as they explore ways to improve the robustness of deep learning models to analog or digital noise using various methods. In this paper, the effect of batch normalization on the robustness property of deep learning models in the presence of analog noise is investigated in order to provide important insights and intuitions on the tradeoff between prediction accuracy and noise resistance of deep learning models. Although this work share some similar visions with that of~\cite{Fagbohungbe2020BenchmarkingIP, Upadhyaya2019ErrorCF} in the modeling of the analog noise and in adding noise to all the layers of the model, this work is unique as it proposes new ways to design and train deep learning model with a controlled tradeoff between model performance and model robustness to analog noise. This work is also different from~\cite{Qiao2019MicroBatchTW,QinBitFlipping} where the noise is modeled as digital noise. Furthermore, it is also different from \cite{Santurkar2018HowDB,Bjorck2018UnderstandingBN} which aim to provide an alternate reasons and mathematical derivation why batch normalization accelerates training. 

\section{{Conclusion}}
\label{sec:conclusion}
The effect of batch normalization on the robustness of deep learning models inference performance to analog noise is investigated in this paper. The investigation is done by comparing the performance of a deep learning model with batch normalization layer with the same deep learning model without batch normalization layer in the presence of noise. The effect of the noise on the model is modeled as a form of weight change. 

This paper established that batch normalization layer negatively impacts on the robustness properties of deep learning model to analog noise. The influence of the batch normalization layer on the noise resistant property of model is such that it increases with increase in the number of batch normalization layer. In fact, the extra performance improvement in model inference due to the presence of the batch normalization layer is lost and the model performs poorly when compared with models without batch normalization layer. In cases where the model training is impossible without batch normalization layer, it is proposed that the minimum amount of batch normalization layer needed to get the model trained and achieved the target model performance is encouraged. This case is defined as partial case of batch normalization. These training paradigm ensures that a tradeoff is achieved between model performance and model's robustness property to analog noise. This paradigm also applies in situation when there is significant difference in inference performance between model with batch normalization layer and model without any batch normalization layer.  It is also observed that for a fixed model, the noise resistant ability of a model is negatively impacted as the complexity of the classification task increases.

\section{Acknowledgments}
\label{acknowledgement}
This research work is supported by the U.S. Office of the Under Secretary of Defense for Research and Engineering (OUSD(R\&E)) under agreement number FA8750-15-2-0119. The U.S. Government is authorized to reproduce and distribute reprints for governmental purposes notwithstanding any copyright notation thereon. The views and conclusions contained herein are those of the authors and should not be interpreted as necessarily representing the official policies or endorsements, either expressed or implied, of the Office of the Under Secretary of Defense for Research and Engineering (OUSD(R\&E)) or the U.S. Government.

\bibliographystyle{IEEEtran}

\bibliography{BatchNormalisation}

\begin{thebibliography}{10}
\providecommand{\url}[1]{#1}
\csname url@samestyle\endcsname
\providecommand{\newblock}{\relax}
\providecommand{\bibinfo}[2]{#2}
\providecommand{\BIBentrySTDinterwordspacing}{\spaceskip=0pt\relax}
\providecommand{\BIBentryALTinterwordstretchfactor}{4}
\providecommand{\BIBentryALTinterwordspacing}{\spaceskip=\fontdimen2\font plus
\BIBentryALTinterwordstretchfactor\fontdimen3\font minus
  \fontdimen4\font\relax}
\providecommand{\BIBforeignlanguage}[2]{{%
\expandafter\ifx\csname l@#1\endcsname\relax
\typeout{** WARNING: IEEEtran.bst: No hyphenation pattern has been}%
\typeout{** loaded for the language `#1'. Using the pattern for}%
\typeout{** the default language instead.}%
\else
\language=\csname l@#1\endcsname
\fi
#2}}
\providecommand{\BIBdecl}{\relax}
\BIBdecl

\bibitem{mixedsignal}
\BIBentryALTinterwordspacing
M.~Klachko, M.~R. Mahmoodi, and D.~B. Strukov, ``Improving noise tolerance of
  mixed-signal neural networks,'' \emph{CoRR}, vol. abs / 1904.01705, 2019.
  [Online]. Available: \url{http://arxiv.org/abs/1904.01705}
\BIBentrySTDinterwordspacing

\bibitem{Li}
H.~{Li}, M.~{Bhargav}, P.~N. {Whatmough}, and H.~. {Philip Wong}, ``On-chip
  memory technology design space explorations for mobile deep neural network
  accelerators,'' in \emph{2019 56th ACM/IEEE Design Automation Conference
  (DAC)}, 2019, pp. 1--6.

\bibitem{joshi}
\BIBentryALTinterwordspacing
V.~Joshi, M.~L. Gallo, I.~Boybat, S.~Haefeli, C.~Piveteau, M.~Dazzi,
  B.~Rajendran, A.~Sebastian, and E.~Eleftheriou, ``Accurate deep neural
  network inference using computational phase-change memory,'' \emph{CoRR},
  vol. abs / 1906.03138, 2019. [Online]. Available:
  \url{http://arxiv.org/abs/1906.03138}
\BIBentrySTDinterwordspacing

\bibitem{charan}
G.~{Charan}, A.~{Mohanty}, X.~{Du}, G.~{Krishnan}, R.~V. {Joshi}, and Y.~{Cao},
  ``Accurate inference with inaccurate rram devices: A joint algorithm-design
  solution,'' \emph{IEEE Journal on Exploratory Solid-State Computational
  Devices and Circuits}, vol.~6, no.~1, pp. 27--35, 2020.

\bibitem{NoisyNN}
C.~Zhou, P.~Kadambi, M.~Mattina, and P.~N. Whatmough, ``Noisy machines:
  Understanding noisy neural networks and enhancing robustness to analog
  hardware errors using distillation,'' \emph{ArXiv}, vol. abs/2001.04974,
  2020.

\bibitem{tpu}
\BIBentryALTinterwordspacing
N.~P. Jouppi, C.~Young, N.~Patil, D.~Patterson, G.~Agrawal, R.~Bajwa, S.~Bates,
  S.~Bhatia, N.~Boden, A.~Borchers, R.~Boyle, P.-l. Cantin, C.~Chao, C.~Clark,
  J.~Coriell, M.~Daley, M.~Dau, J.~Dean, B.~Gelb, T.~V. Ghaemmaghami,
  R.~Gottipati, W.~Gulland, R.~Hagmann, C.~R. Ho, D.~Hogberg, J.~Hu, R.~Hundt,
  D.~Hurt, J.~Ibarz, A.~Jaffey, A.~Jaworski, A.~Kaplan, H.~Khaitan,
  D.~Killebrew, A.~Koch, N.~Kumar, S.~Lacy, J.~Laudon, J.~Law, D.~Le, C.~Leary,
  Z.~Liu, K.~Lucke, A.~Lundin, G.~MacKean, A.~Maggiore, M.~Mahony, K.~Miller,
  R.~Nagarajan, R.~Narayanaswami, R.~Ni, K.~Nix, T.~Norrie, M.~Omernick,
  N.~Penukonda, A.~Phelps, J.~Ross, M.~Ross, A.~Salek, E.~Samadiani, C.~Severn,
  G.~Sizikov, M.~Snelham, J.~Souter, D.~Steinberg, A.~Swing, M.~Tan,
  G.~Thorson, B.~Tian, H.~Toma, E.~Tuttle, V.~Vasudevan, R.~Walter, W.~Wang,
  E.~Wilcox, and D.~H. Yoon, ``In-datacenter performance analysis of a tensor
  processing unit,'' \emph{SIGARCH Comput. Archit. News}, vol.~45, no.~2, p.
  1–12, jun 2017. [Online]. Available:
  \url{https://doi.org/10.1145/3140659.3080246}
\BIBentrySTDinterwordspacing

\bibitem{gpu}
A.~Maghazeh, U.~D. Bordoloi, P.~Eles, and Z.~Peng, ``General purpose computing
  on low-power embedded gpus: Has it come of age?'' in \emph{2013 International
  Conference on Embedded Computer Systems: Architectures, Modeling, and
  Simulation (SAMOS)}, 2013, pp. 1--10.

\bibitem{han2015deep}
S.~Han, H.~Mao, and W.~J. Dally, ``Deep compression: Compressing deep neural
  networks with pruning, trained quantization and huffman coding,'' \emph{arXiv
  preprint arXiv:1510.00149}, 2015.

\bibitem{luo2017thinet}
J.-H. Luo, J.~Wu, and W.~Lin, ``Thinet: A filter level pruning method for deep
  neural network compression,'' in \emph{Proceedings of the IEEE international
  conference on computer vision}, 2017, pp. 5058--5066.

\bibitem{hinton2015distilling}
G.~Hinton, O.~Vinyals, and J.~Dean, ``Distilling the knowledge in a neural
  network,'' \emph{arXiv preprint arXiv:1503.02531}, 2015.

\bibitem{xiao}
T.~P. Xiao, C.~H. Bennett, B.~Feinberg, S.~Agarwal, and M.~J. Marinella,
  ``Analog architectures for neural network acceleration based on non-volatile
  memory,'' \emph{Applied Physics Reviews}, vol.~7, no.~3, 9 2020.

\bibitem{MahmoodiAnalog}
\BIBentryALTinterwordspacing
M.~R. Mahmoodi and D.~Strukov, ``An ultra-low energy internally analog,
  externally digital vector-matrix multiplier based on nor flash memory
  technology,'' in \emph{Proceedings of the 55th Annual Design Automation
  Conference}, ser. DAC '18.\hskip 1em plus 0.5em minus 0.4em\relax New York,
  NY, USA: Association for Computing Machinery, 2018. [Online]. Available:
  \url{https://doi.org/10.1145/3195970.3195989}
\BIBentrySTDinterwordspacing

\bibitem{Ni}
L.~{Ni}, Z.~{Liu}, H.~{Yu}, and R.~V. {Joshi}, ``An energy-efficient digital
  reram-crossbar-based cnn with bitwise parallelism,'' \emph{IEEE Journal on
  Exploratory Solid-State Computational Devices and Circuits}, vol.~3, pp.
  37--46, 2017.

\bibitem{Shen_2017}
\BIBentryALTinterwordspacing
Y.~Shen, N.~C. Harris, S.~Skirlo, M.~Prabhu, T.~Baehr-Jones, M.~Hochberg,
  X.~Sun, S.~Zhao, H.~Larochelle, D.~Englund, and et~al., ``Deep learning with
  coherent nanophotonic circuits,'' \emph{Nature Photonics}, vol.~11, no.~7, p.
  441–446, Jun 2017. [Online]. Available:
  \url{http://dx.doi.org/10.1038/nphoton.2017.93}
\BIBentrySTDinterwordspacing

\bibitem{Bennett2020}
\BIBentryALTinterwordspacing
C.~H. Bennett, T.~P. Xiao, R.~Dellana, B.~Feinberg, S.~Agarwal, M.~J.
  Marinella, V.~Agrawal, V.~Prabhakar, K.~Ramkumar, L.~Hinh, and et~al.,
  ``Device-aware inference operations in sonos nonvolatile memory arrays,''
  \emph{2020 IEEE International Reliability Physics Symposium (IRPS)}, Apr
  2020. [Online]. Available:
  \url{http://dx.doi.org/10.1109/IRPS45951.2020.9129313}
\BIBentrySTDinterwordspacing

\bibitem{Burr}
G.~W. Burr, R.~M. Shelby, A.~Sebastian, S.~Kim, S.~Kim, S.~Sidler, K.~Virwani,
  M.~Ishii, P.~Narayanan, A.~Fumarola, L.~L. Sanches, I.~Boybat, M.~L. Gallo,
  K.~Moon, J.~Woo, H.~Hwang, and Y.~Leblebici, ``Neuromorphic computing using
  non-volatile memory,'' \emph{Advances in Physics: X}, vol.~2, no.~1, pp.
  89--124, 2017.

\bibitem{marinella}
M.~J. {Marinella}, S.~{Agarwal}, A.~{Hsia}, I.~{Richter}, R.~{Jacobs-Gedrim},
  J.~{Niroula}, S.~J. {Plimpton}, E.~{Ipek}, and C.~D. {James}, ``Multiscale
  co-design analysis of energy, latency, area, and accuracy of a reram analog
  neural training accelerator,'' \emph{IEEE Journal on Emerging and Selected
  Topics in Circuits and Systems}, vol.~8, no.~1, pp. 86--101, 2018.

\bibitem{MITTAL2020101689}
\BIBentryALTinterwordspacing
S.~Mittal, ``A survey on modeling and improving reliability of dnn algorithms
  and accelerators,'' \emph{Journal of Systems Architecture}, vol. 104, p.
  101689, 2020. [Online]. Available:
  \url{http://www.sciencedirect.com/science/article/pii/S1383762119304965}
\BIBentrySTDinterwordspacing

\bibitem{Fagbohungbe2020BenchmarkingIP}
O.~I. Fagbohungbe and L.~Qian, ``Benchmarking inference performance of deep
  learning models on analog devices,'' \emph{ArXiv}, vol. abs/2011.11840, 2020.

\bibitem{Merolla2016DeepNN}
P.~Merolla, R.~Appuswamy, J.~Arthur, S.~K. Esser, and D.~Modha, ``Deep neural
  networks are robust to weight binarization and other non-linear
  distortions,'' \emph{ArXiv}, vol. abs/1606.01981, 2016.

\bibitem{Ioffe2015BatchNA}
S.~Ioffe and C.~Szegedy, ``Batch normalization: Accelerating deep network
  training by reducing internal covariate shift,'' \emph{ArXiv}, vol.
  abs/1502.03167, 2015.

\bibitem{Ioffe2017BatchRT}
S.~Ioffe, ``Batch renormalization: Towards reducing minibatch dependence in
  batch-normalized models,'' in \emph{NIPS}, 2017.

\bibitem{Wu2019L1B}
S.~Wu, G.~Li, L.~Deng, L.~Liu, D.~Wu, Y.~Xie, and L.~Shi, ``\$l1\$ -norm batch
  normalization for efficient training of deep neural networks,'' \emph{IEEE
  Transactions on Neural Networks and Learning Systems}, vol.~30, pp.
  2043--2051, 2019.

\bibitem{Santurkar2018HowDB}
S.~Santurkar, D.~Tsipras, A.~Ilyas, and A.~Madry, ``How does batch
  normalization help optimization?'' in \emph{NeurIPS}, 2018.

\bibitem{Qiao2019MicroBatchTW}
S.~Qiao, H.~Wang, C.~Liu, W.~Shen, and A.~Yuille, ``Micro-batch training with
  batch-channel normalization and weight standardization,'' 2019.

\bibitem{Bjorck2018UnderstandingBN}
J.~Bjorck, C.~P. Gomes, and B.~Selman, ``Understanding batch normalization,''
  in \emph{NeurIPS}, 2018.

\bibitem{zhou2019toward}
M.~Zhou, T.~Liu, Y.~Li, D.~Lin, E.~Zhou, and T.~Zhao, ``Toward understanding
  the importance of noise in training neural networks,'' in \emph{International
  Conference on Machine Learning}.\hskip 1em plus 0.5em minus 0.4em\relax PMLR,
  2019, pp. 7594--7602.

\bibitem{wen2018smoothout}
W.~Wen, Y.~Wang, F.~Yan, C.~Xu, C.~Wu, Y.~Chen, and H.~Li, ``Smoothout:
  Smoothing out sharp minima to improve generalization in deep learning,''
  \emph{arXiv preprint arXiv:1805.07898}, 2018.

\bibitem{resnet}
\BIBentryALTinterwordspacing
K.~He, X.~Zhang, S.~Ren, and J.~Sun, ``Deep residual learning for image
  recognition,'' \emph{CoRR}, vol. abs/1512.03385, 2015. [Online]. Available:
  \url{http://arxiv.org/abs/1512.03385}
\BIBentrySTDinterwordspacing

\bibitem{vgg}
K.~Simonyan and A.~Zisserman, ``Very deep convolutional networks for
  large-scale image recognition,'' \emph{CoRR}, vol. abs/1409.1556, 2015.

\bibitem{Wu2018GroupN}
Y.~Wu and K.~He, ``Group normalization,'' in \emph{ECCV}, 2018.

\bibitem{Ba2016LayerN}
J.~Ba, J.~Kiros, and G.~E. Hinton, ``Layer normalization,'' \emph{ArXiv}, vol.
  abs/1607.06450, 2016.

\bibitem{Salimans2016WeightNA}
T.~Salimans and D.~P. Kingma, ``Weight normalization: A simple
  reparameterization to accelerate training of deep neural networks,'' in
  \emph{NIPS}, 2016.

\bibitem{SIETSMA199167}
\BIBentryALTinterwordspacing
J.~Sietsma and R.~J. Dow, ``Creating artificial neural networks that
  generalize,'' \emph{Neural Networks}, vol.~4, no.~1, pp. 67--79, 1991.
  [Online]. Available:
  \url{https://www.sciencedirect.com/science/article/pii/0893608091900332}
\BIBentrySTDinterwordspacing

\bibitem{Minnix}
J.~I. {Minnix}, ``Fault tolerance of the backpropagation neural network trained
  on noisy inputs,'' in \emph{[Proceedings 1992] IJCNN International Joint
  Conference on Neural Networks}, vol.~1, 1992, pp. 847--852 vol.1.

\bibitem{Meng}
X.~{Meng}, C.~{Liu}, Z.~{Zhang}, and D.~{Wang}, ``Noisy training for deep
  neural networks,'' in \emph{2014 IEEE China Summit International Conference
  on Signal and Information Processing (ChinaSIP)}, 2014, pp. 16--20.

\bibitem{Rusiecki2014TrainingNN}
A.~Rusiecki, M.~Kordos, T.~Kaminski, and K.~Gren, ``Training neural networks on
  noisy data,'' in \emph{ICAISC}, 2014.

\bibitem{Murray}
A.~F. {Murray} and P.~J. {Edwards}, ``Enhanced mlp performance and fault
  tolerance resulting from synaptic weight noise during training,'' \emph{IEEE
  Transactions on Neural Networks}, vol.~5, no.~5, pp. 792--802, 1994.

\bibitem{Qin}
\BIBentryALTinterwordspacing
M.~Qin and D.~Vucinic, ``Noisy computations during inference: Harmful or
  helpful?'' \emph{CoRR}, vol. abs / 1811.10649, 2018. [Online]. Available:
  \url{http://arxiv.org/abs/1811.10649}
\BIBentrySTDinterwordspacing

\bibitem{Qin2018TrainingRN}
Q.~Minghai and V.~D., ``Training recurrent neural networks against noisy
  computations during inference,'' \emph{2018 52nd Asilomar Conference on
  Signals, Systems, and Computers}, pp. 71--75, 2018.

\bibitem{Miyashita}
D.~{Miyashita}, S.~{Kousai}, T.~{Suzuki}, and J.~{Deguchi}, ``A neuromorphic
  chip optimized for deep learning and cmos technology with time-domain analog
  and digital mixed-signal processing,'' \emph{IEEE Journal of Solid-State
  Circuits}, vol.~52, no.~10, pp. 2679--2689, 2017.

\bibitem{An}
\BIBentryALTinterwordspacing
G.~An, ``The effects of adding noise during backpropagation training on a
  generalization performance,'' vol.~8, no.~3, p. 643–674, Apr. 1996.
  [Online]. Available: \url{https://doi.org/10.1162/neco.1996.8.3.643}
\BIBentrySTDinterwordspacing

\bibitem{Bishop}
C.~M. {Bishop}, ``Training with noise is equivalent to tikhonov
  regularization,'' \emph{Neural Computation}, vol.~7, no.~1, pp. 108--116,
  1995.

\bibitem{Matsuoka}
K.~{Matsuoka}, ``Noise injection into inputs in back-propagation learning,''
  \emph{IEEE Transactions on Systems, Man, and Cybernetics}, vol.~22, no.~3,
  pp. 436--440, 1992.

\bibitem{Holmstrom}
L.~{Holmstrom} and P.~{Koistinen}, ``Using additive noise in back-propagation
  training,'' \emph{IEEE Transactions on Neural Networks}, vol.~3, no.~1, pp.
  24--38, 1992.

\bibitem{noh2017regularizing}
H.~Noh, T.~You, J.~Mun, and B.~Han, ``Regularizing deep neural networks by
  noise: Its interpretation and optimization,'' 2017.

\bibitem{Zheng2016ImprovingTR}
S.~Zheng, Y.~Song, T.~Leung, and I.~J. Goodfellow, ``Improving the robustness
  of deep neural networks via stability training,'' \emph{2016 IEEE Conference
  on Computer Vision and Pattern Recognition (CVPR)}, pp. 4480--4488, 2016.

\bibitem{Chen2019UnderstandingAU}
P.~Chen, B.~Liao, G.~Chen, and S.~Zhang, ``Understanding and utilizing deep
  neural networks trained with noisy labels,'' in \emph{ICML}, 2019.

\bibitem{Reed2015TrainingDN}
S.~Reed, H.~Lee, D.~Anguelov, C.~Szegedy, D.~Erhan, and A.~Rabinovich,
  ``Training deep neural networks on noisy labels with bootstrapping,''
  \emph{CoRR}, vol. abs/1412.6596, 2015.

\bibitem{Bo}
G.~M. {Bo}, D.~D. {Caviglia}, and M.~{Valle}, ``An on-chip learning neural
  network,'' in \emph{Proceedings of the IEEE-INNS-ENNS International Joint
  Conference on Neural Networks. IJCNN 2000. Neural Computing: New Challenges
  and Perspectives for the New Millennium}, vol.~4, 2000, pp. 66--71 vol.4.

\bibitem{schmid}
A.~Schmid, Y.~Leblebici, and D.~Mlynek, ``Mixed analogue-digital
  artificial-neural-network architecture with on-chip learning,''
  \emph{Circuits, Devices and Systems, IEE Proceedings -}, vol. 146, pp. 345 --
  349, 01 2000.

\bibitem{Bayraktaroglu}
I.~{Bayraktaroglu}, A.~S. {Ogrenci}, G.~{Dundar}, S.~{Balkir}, and
  E.~{Alpaydin}, ``Annsys (an analog neural network synthesis system),'' in
  \emph{Proceedings of International Conference on Neural Networks (ICNN'97)},
  vol.~2, 1997, pp. 910--915 vol.2.

\bibitem{NeuroSmith}
\BIBentryALTinterwordspacing
S.~Schmitt, J.~Klaehn, G.~Bellec, A.~Gr{\"{u}}bl, M.~Guettler, A.~Hartel,
  S.~Hartmann, D.~H. de~Oliveira, K.~Husmann, V.~Karasenko, M.~Kleider,
  C.~Koke, C.~Mauch, E.~M{\"{u}}ller, P.~M{\"{u}}ller, J.~Partzsch, M.~A.
  Petrovici, S.~Schiefer, S.~Scholze, B.~Vogginger, R.~A. Legenstein, W.~Maass,
  C.~Mayr, J.~Schemmel, and K.~Meier, ``Neuromorphic hardware in the loop:
  Training a deep spiking network on the brainscales wafer-scale system,''
  \emph{CoRR}, vol. abs/1703.01909, 2017. [Online]. Available:
  \url{http://arxiv.org/abs/1703.01909}
\BIBentrySTDinterwordspacing

\bibitem{Garg2021DynamicPA}
S.~Garg, J.~Lou, A.~Jain, and M.~Nahmias, ``Dynamic precision analog computing
  for neural networks,'' \emph{ArXiv}, vol. abs/2102.06365, 2021.

\bibitem{Upadhyaya2019ErrorCF}
P.~Upadhyaya, X.~Yu, J.~Mink, J.~Cordero, P.~Parmar, and A.~Jiang, ``Error
  correction for hardware-implemented deep neural networks,'' 2019.

\bibitem{Upadhyaya2019ErrorCF2}
P.~Upadhyaya, X.~Yu, J.~Mink, C.~J., P.~Parmar, and A.~Jiang, ``Error
  correction for noisy neural networks,'' 10 2019.

\bibitem{QinBitFlipping}
M.~Qin, C.~Sun, and D.~Vucinic, ``Improving robustness of neural networks
  against bit flipping errors during inference,'' \emph{Journal of Image and
  Graphics}, vol.~6, pp. 181--186, 01 2018.

\bibitem{huang2020functional}
K.~Huang, P.~Siegel, and A.~Jiang, ``Functional error correction for robust
  neural networks,'' 2020.

\bibitem{LiuDAC2019}
\BIBentryALTinterwordspacing
T.~Liu, W.~Wen, L.~Jiang, Y.~Wang, C.~Yang, and G.~Quan, ``A fault-tolerant
  neural network architecture,'' in \emph{Proceedings of the 56th Annual Design
  Automation Conference 2019}, ser. DAC '19.\hskip 1em plus 0.5em minus
  0.4em\relax New York, NY, USA: Association for Computing Machinery, 2019.
  [Online]. Available: \url{https://doi.org/10.1145/3316781.3317742}
\BIBentrySTDinterwordspacing

\bibitem{Zhang2019}
J.~J. Zhang, K.~Basu, and S.~Garg, ``Fault-tolerant systolic array based
  accelerators for deep neural network execution,'' \emph{IEEE Design Test},
  vol.~36, no.~5, pp. 44--53, 2019.

\bibitem{ShieldeNN}
N.~Khoshavi, A.~Roohi, C.~Broyles, S.~Sargolzaei, Y.~Bi, and D.~Z. Pan,
  ``Shieldenn: Online accelerated framework for fault-tolerant deep neural
  network architectures,'' in \emph{2020 57th ACM/IEEE Design Automation
  Conference (DAC)}, 2020, pp. 1--6.

\end{thebibliography}

$ $ \\

\begin{IEEEbiography}[{\includegraphics[width=1in,height=1.25in,clip,keepaspectratio]{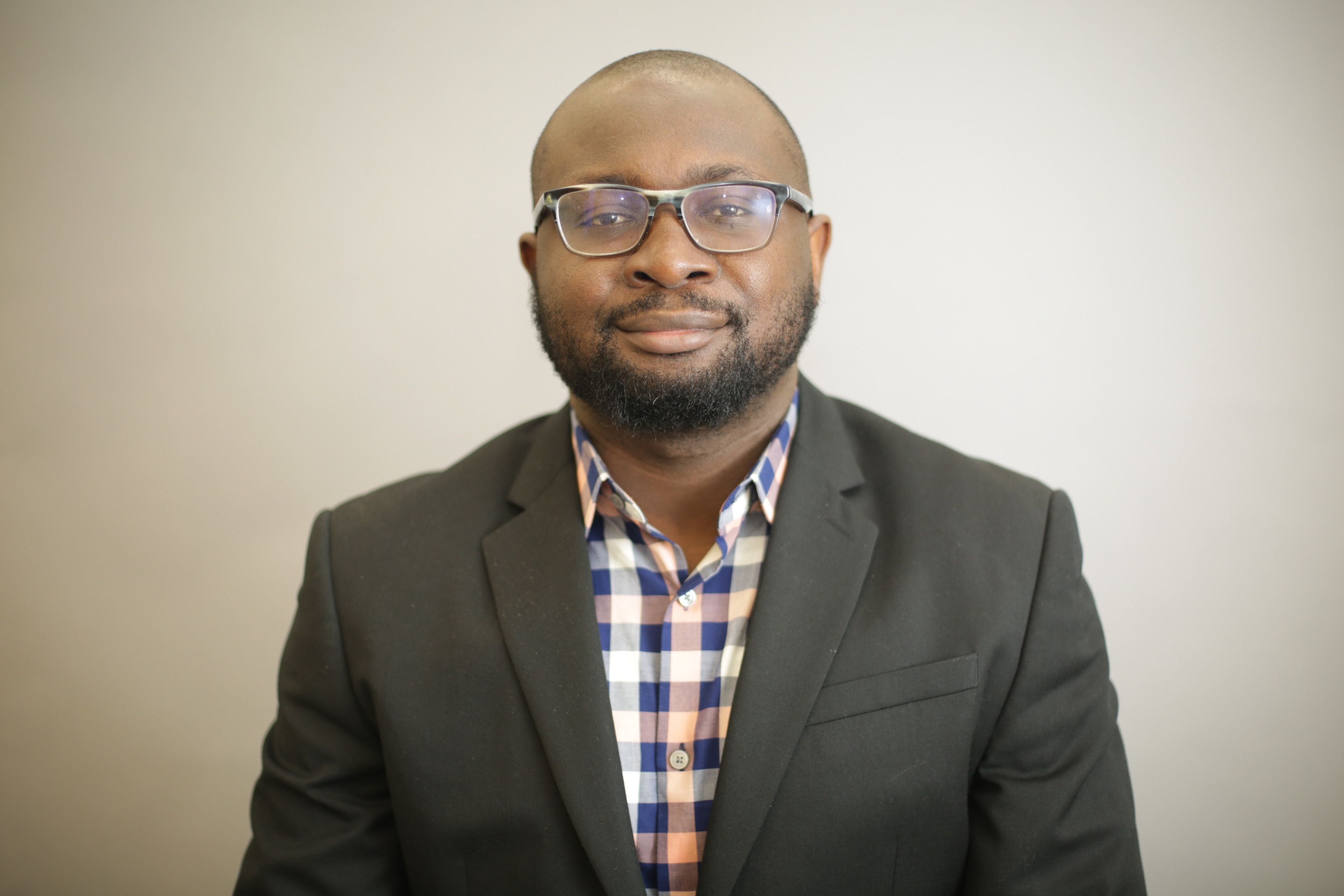}}]{Omobayode Fagbohungbe} is currently working towards his Ph.D. degree at U.S. DOD Center of Excellence in Research and Education for Big Military Data Intelligence (CREDIT Center),  Department of Electrical and Computer Engineering, Prairie View A\&M University, Texas, USA.  Prior to now, he received the B.S. degree in Electronic and Electrical Engineering from Obafemi Awolowo University, Ile-Ife, Nigeria and the M.S. degree in Control Engineering from the University of Manchester, Manchester, United Kingdom. His research interests are in the area of big data, data science, robust deep learning models and artificial intelligence.
\end{IEEEbiography}

%

\begin{IEEEbiography}[{\includegraphics[width=1in,height=1.25in,clip,keepaspectratio]{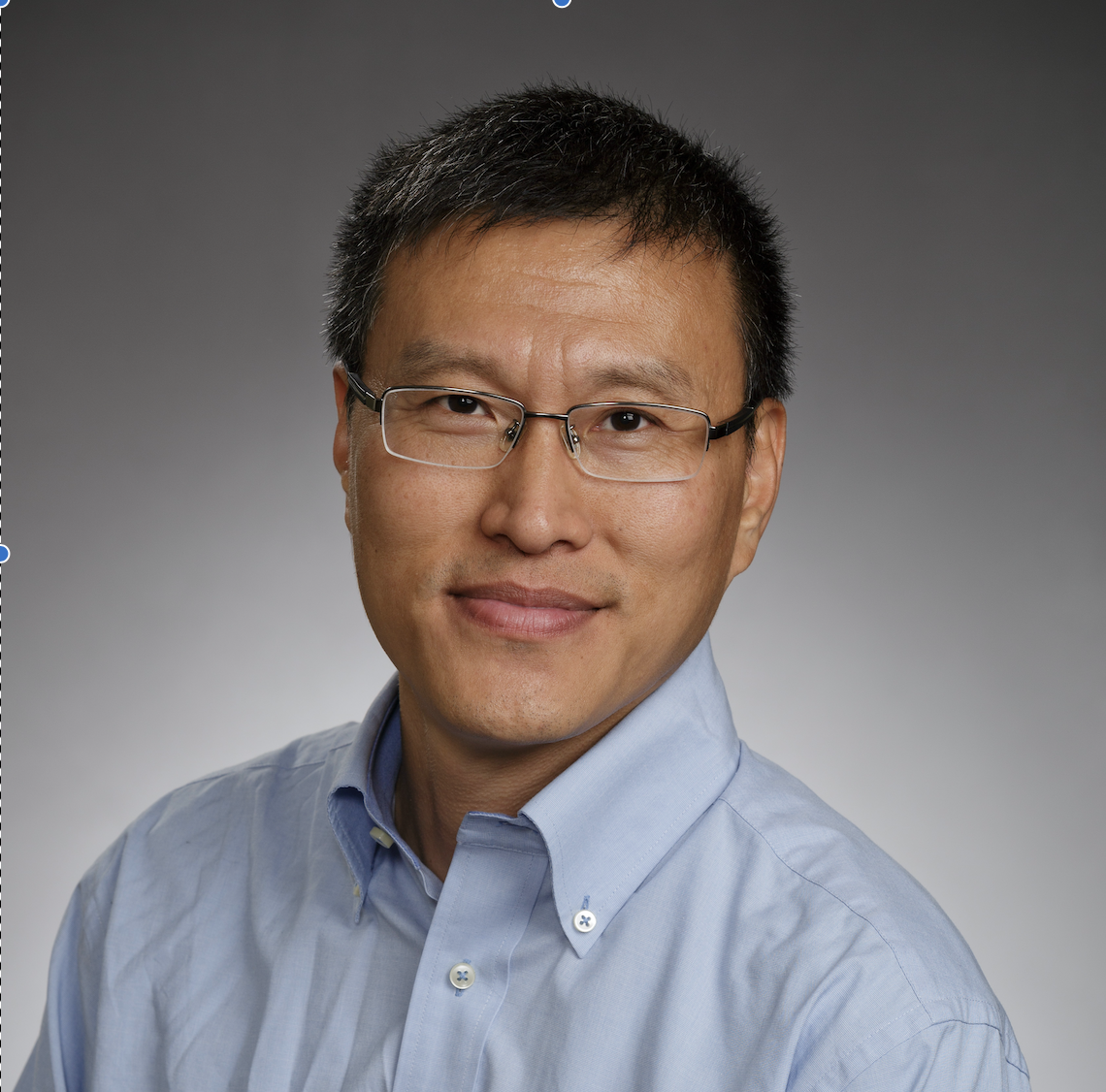}}]{Lijun Qian} (SM'08) is Regents Professor and holds the AT\&T Endowment in the Department of Electrical and Computer Engineering at Prairie View A\&M University (PVAMU), a member of the Texas A\&M University System, Prairie View, Texas, USA. He is also the Director of the Center of Excellence in Research and Education for Big Military Data Intelligence (CREDIT Center). He received BS from Tsinghua University, MS from Technion-Israel Institute of Technology, and PhD from Rutgers University. Before joining PVAMU, he was a member of technical staff of Bell-Labs Research at Murray Hill, New Jersey. He was a visiting professor of Aalto University, Finland. His research interests are in the area of big data processing, artificial intelligence, wireless communications and mobile networks, network security and intrusion detection, and computational and systems biology.
\end{IEEEbiography}


\end{document}